\ifdefined\XeTeXversion
\else
  \pdfoutput=1
\fi
\documentclass[11pt]{article}

\usepackage[preprint]{acl}
\usepackage{times}
\usepackage{latexsym}
\usepackage[T1]{fontenc}
\usepackage[utf8]{inputenc}
\usepackage{microtype}
\usepackage{inconsolata}
\usepackage{amsmath,amssymb}
\usepackage{booktabs}
\usepackage{xspace}
\usepackage{graphicx}
\usepackage{xcolor}
\usepackage[section]{placeins}
\usepackage{float}
\usepackage{enumitem}

\floatstyle{ruled}
\newfloat{algorithm}{t}{loa}
\floatname{algorithm}{Algorithm}


\newcommand{\method}{\leavevmode{\fontfamily{lmtt}\selectfont \textbf{TRIAL}}\xspace}

\title{Trajectory-Relative Hindsight Distillation for Agentic Reinforcement Learning}

\author{
Haoyu Zheng\textsuperscript{1,3}\thanks{Work done during internship at Tencent.},
Yun Zhu\textsuperscript{2},
Qing Wang\textsuperscript{3},
Wenqiao Zhang\textsuperscript{1}\thanks{Corresponding author.}
\\
\textsuperscript{1}Zhejiang University
\quad \textsuperscript{2}Shanghai AI Laboratory
\quad \textsuperscript{3}Tencent
}

\begin{document}
\maketitle
\begin{abstract}
Recent agentic reinforcement learning methods use hindsight to complement sparse outcome rewards. However, a completed rollout can yield many such signals, leaving their appropriate allocation across turns unclear. We introduce \method, a trajectory-relative hindsight distillation framework with a unified turn-aligned scoring protocol. For each decision turn, \method extracts an outcome view of that decision's realized consequence and evaluates the same response under ordinary and hindsight-conditioned contexts. The signed log-probability gap determines the direction and local strength of token-level supervision, while turn-level magnitudes are normalized jointly over the realized trajectory. The resulting allocation multipliers have an eligible-token-weighted mean of one, redistributing dense supervision across turns while fixing its average multiplier. Experiments on WebShop and ALFWorld with different backbones show that \method outperforms GRPO across all eight combinations of backbone, environment, and evaluation metric, while achieving the best or tied-best performance among six methods on six of them. On WebShop with Qwen3-1.7B, \method improves the success rate from 56.4\% to 75.2\% and the task score from 78.7\% to 85.7\%. Controlled ablations further show that trajectory-relative turn allocation provides substantial gains beyond those of dense hindsight distillation alone.
\end{abstract}

\section{Introduction}
\label{sec:introduction}

Agentic reinforcement learning commonly optimizes sparse outcome rewards.
GRPO, for example, compares complete rollouts for the same task using relative
outcome signals
\citep{shao2024deepseekmathpushinglimitsmathematical}.
A multi-turn rollout, however, contains many decisions made under changing
observations, and its final outcome does not specify how supervision should be
distributed among them. Existing work reduces this granularity mismatch
through finer rollout or step-level groupings
\citep{wang2025ragenunderstandingselfevolutionllm,
feng2025groupingrouppolicyoptimizationllm},
but remains centered on estimating outcome-based advantages.

Hindsight distillation complements sparse outcome rewards with dense
supervision by reevaluating realized actions with their turn-aligned realized
consequences
\citep{agarwal2024onpolicydistillationlanguagemodels,
zhao2026selfdistilledreasoneronpolicyselfdistillation}.
Recent methods use hindsight-induced policy discrepancies to modulate token
updates, distribute supervision within steps, or select turns for feedback
\citep{yang2026selfdistilledrlvr,
lu2026selfdistilledagenticreinforcementlearning,
li2026distillselectivehindsightdistillation,
zhang2026stepopsdstepawareonlinepreference,
yeo2026hintsdtargetedhindsightselfdistillation}.
However, a completed rollout can produce discrepancies at many turns. Each
signed token gap encodes a local revision, but its raw magnitude does not state
how strongly that turn should be emphasized relative to other turns in the
same interaction. The resulting problem is how to
\textbf{preserve signed token-level revision while redistributing hindsight
supervision across turns} under a calibrated allocation profile.

We introduce \method{},\footnote{Code is available at
\url{https://github.com/Chihaya-Anon-chan/TRIAL}.} a trajectory-relative
hindsight distillation framework
with one shared protocol across interactive environments. For every decision
turn, an outcome view exposes the locally realized consequence of that decision
to a training-time scorer, which reevaluates the same realized response. The
resulting signed token-level log-probability gap determines update direction and
local strength. To allocate this supervision across turns, \method{} aggregates
the absolute gaps within each turn and compares the turn's share of total
revision magnitude with its share of eligible tokens. This produces a
trajectory-relative profile whose turn multipliers have an
eligible-token-weighted mean of one, emphasizing turns with stronger revision
evidence per token while keeping the average allocation multiplier fixed.
\looseness=-1

\begin{figure}[t]
  \centering
  \includegraphics[width=\columnwidth]{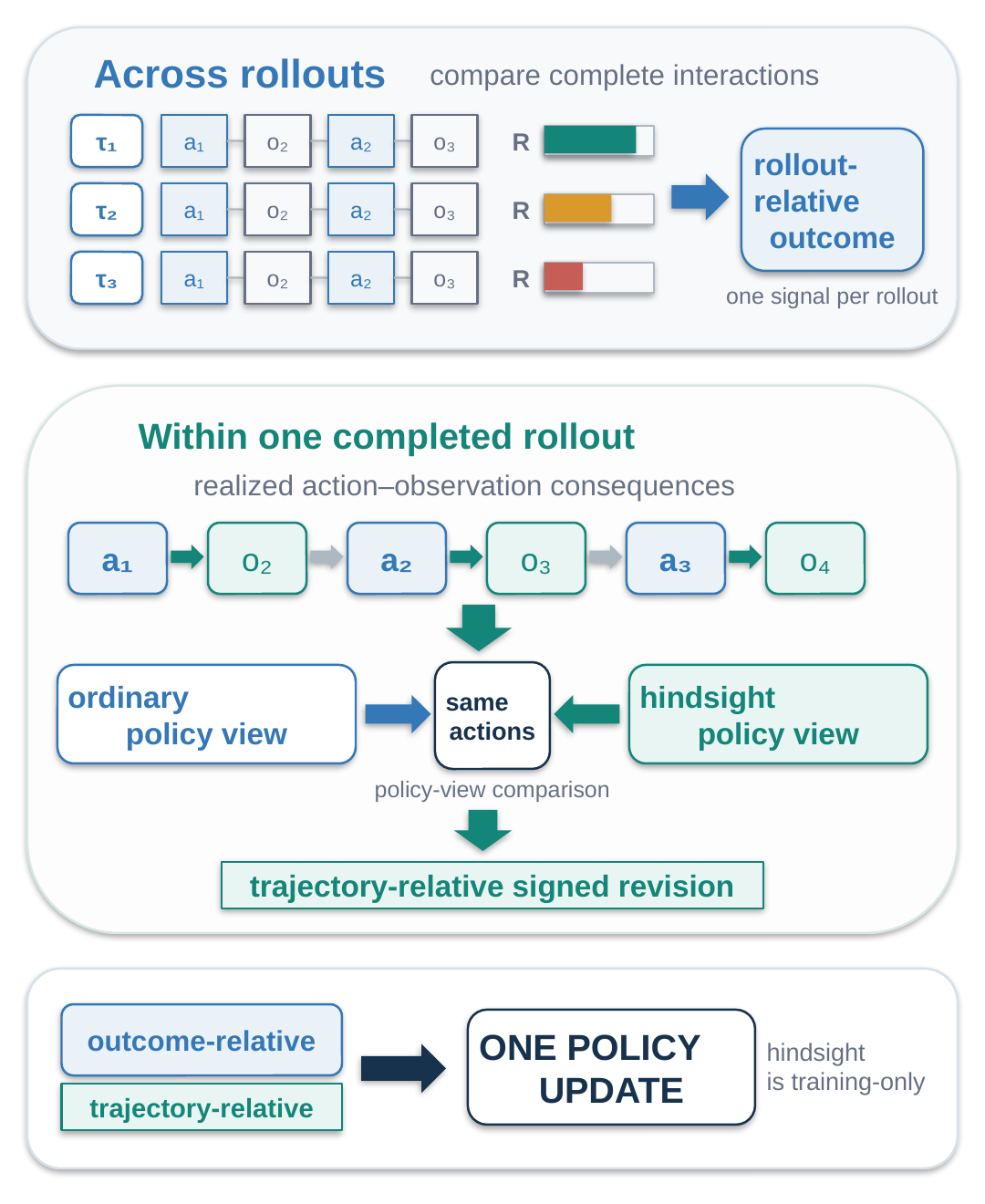}
  \caption{Complementary feedback granularities in a multi-turn agent.
  Outcome feedback compares complete trajectories. At each decision turn, a
  turn-aligned outcome view exposes the realized action--observation consequence
  ($a\!\rightarrow\!o$) in a hindsight-conditioned context. Comparing the ordinary and hindsight
  contexts on the same realized tokens yields signed token-level supervision,
  whose magnitude is normalized across turns to produce a
  trajectory-relative allocation profile.}
  \label{fig:intro-overview}
\end{figure}

The resulting profile reweights the signed token gaps to form a dense
hindsight objective. GRPO continues to provide rollout-level outcome
supervision, while \method{} allocates hindsight supervision among tokens and turns within each rollout. The profile normalization keeps the average allocation weight fixed, and a scalar clamp bounds the dense contribution relative to the detached GRPO loss magnitude without changing GRPO's advantage construction. All hindsight contexts, discrepancy scores, and allocation profiles are used only during training, so deployment retains the ordinary online policy without additional inference cost. Figure~\ref{fig:intro-overview} summarizes this decomposition.

We evaluate \method{} on WebShop
\citep{yao2022webshop} and ALFWorld
\citep{shridhar2021alfworldaligningtextembodied} using Qwen2.5-3B and
Qwen3-1.7B. \method{} outperforms GRPO across all eight combinations of
backbone, environment, and evaluation metric, and ranks best or tied best
among six methods in six cases. On WebShop with Qwen3-1.7B, it improves the
success rate from 56.4\% to 75.2\% and the task score from 78.7\% to 85.7\%.
Controlled ablations keep the hindsight pathway fixed while comparing the
trajectory-relative profile with unit and source-shuffled profiles. The
trajectory-relative profile outperforms the unit profile on all four WebShop
and ALFWorld metrics, showing that relative turn allocation provides benefits
beyond dense hindsight distillation alone.

Our contributions are:
\begin{itemize}
  \item We formulate trajectory-relative hindsight allocation as a joint
  calibration problem that couples signed token-level revision with relative
  turn emphasis while controlling the average allocation multiplier.

  \item We introduce \method{}, which derives signed token supervision from
  ordinary and hindsight-conditioned contexts and constructs a
  trajectory-normalized turn profile. Its eligible-token-weighted mean is one,
  so redistribution across turns keeps the average multiplier fixed.

  \item We evaluate \method{} against five baselines on two interactive
  environments with two backbone models. Full-set comparisons and controlled
  ablations demonstrate consistent improvements over GRPO and benefits beyond
  dense hindsight distillation alone.
\end{itemize}

\section{Related Work}
\label{sec:related-work}

\paragraph{Outcome-relative optimization for language agents.}
Instruction-following models have progressed from fine-grained visual
grounding and referring \citep{yuan2024videorefersuite, yuan2025pixelrefer,
yuan2026instructsam} to architectures that adapt a shared backbone across
heterogeneous multimodal tasks and domains \citep{zhang2024hyperllava,
lin2025healthgpt}, and are increasingly deployed as agents that
plan and act over many turns \citep{zheng2026pilot, gao2026visualthinkvla}.
Training such agents from environment outcomes is dominated by group-relative
objectives.
GRPO estimates a critic-free advantage by comparing rewards within a rollout
group \citep{shao2024deepseekmathpushinglimitsmathematical}, but sparse
multi-turn returns obscure which decisions matter. This mismatch compounds as
each action changes later observations while reward often arrives only at
termination. RAGEN analyzes these dynamics
\citep{wang2025ragenunderstandingselfevolutionllm}; GiGPO adds
state-aligned step groups \citep{feng2025groupingrouppolicyoptimizationllm};
and concurrent sibling or counterfactual methods shape token advantages
\citep{ding2026policygradientchargesiblingguided,
meng2026craftcounterfactualcreditassignment}. These methods refine
outcome-relative credit: local structure changes how reward-directed
preference is estimated. \method{} leaves that comparison unchanged and instead
allocates hindsight-induced policy revision among turns after interaction.

\paragraph{On-policy hindsight distillation and fine-grained allocation.}
Hindsight Experience Replay relabels realized experience as additional goals
\citep{andrychowicz2017hindsight}, while asymmetric actor--critic uses
training-only information without changing deployed policy inputs
\citep{pinto2018asymmetric}. On-policy language-model distillation, including
multi-turn variants, learns from student-generated sequences
\citep{agarwal2024onpolicydistillationlanguagemodels,
hubotter2026reinforcementlearningselfdistillation,
wang2026skillsdskillconditionedselfdistillationmultiturn,
zheng2026maigomitigatinglostinconversation}. Our direct baselines use this
signal differently: OPSD scores a privileged same-model view, RLSD and SDAR
couple distillation with reward learning, and SERL selectively distills
multi-turn hindsight
\citep{zhao2026selfdistilledreasoneronpolicyselfdistillation,
yang2026selfdistilledrlvr,lu2026selfdistilledagenticreinforcementlearning,
li2026distillselectivehindsightdistillation}. None of these baselines supplies
a mean-one, trajectory-relative turn profile. TCOD progressively expands the
trajectory depth exposed to the student, while HiSR uses hindsight likelihood
ratios to modulate segment-level process rewards
\citep{wang2026tcodexploringtemporalcurriculum,
lu2026hisrhindsightinformationmodulated}. Other methods reweight tokens or
select particular steps and turns
\citep{xu2026tiptokenimportanceonpolicy,
li2026geargranularityadaptiveadvantagereweighting,
zhang2026stepopsdstepawareonlinepreference,
yeo2026hintsdtargetedhindsightselfdistillation}.
Concurrent work explores turn-aware schedules and selection
\citep{tan2026atodannealedturnawareonpolicy,
zhou2026sageopdselectiveagentguidedintervention,
zhou2026turnopdmakingonpolicydistillation}. \method{} instead couples all
eligible turns in a completed trajectory: same-token hindsight discrepancy
defines revision mass, eligible-token mass provides its reference, and their
ratio yields a mean-one profile that retains signed local revision while
calibrating relative emphasis across turns.

\section{Method}
\label{sec:method}

\begin{figure*}[t]
  \centering
  \includegraphics[
    width=\textwidth,
    trim=53 36 119 129,
    clip
  ]{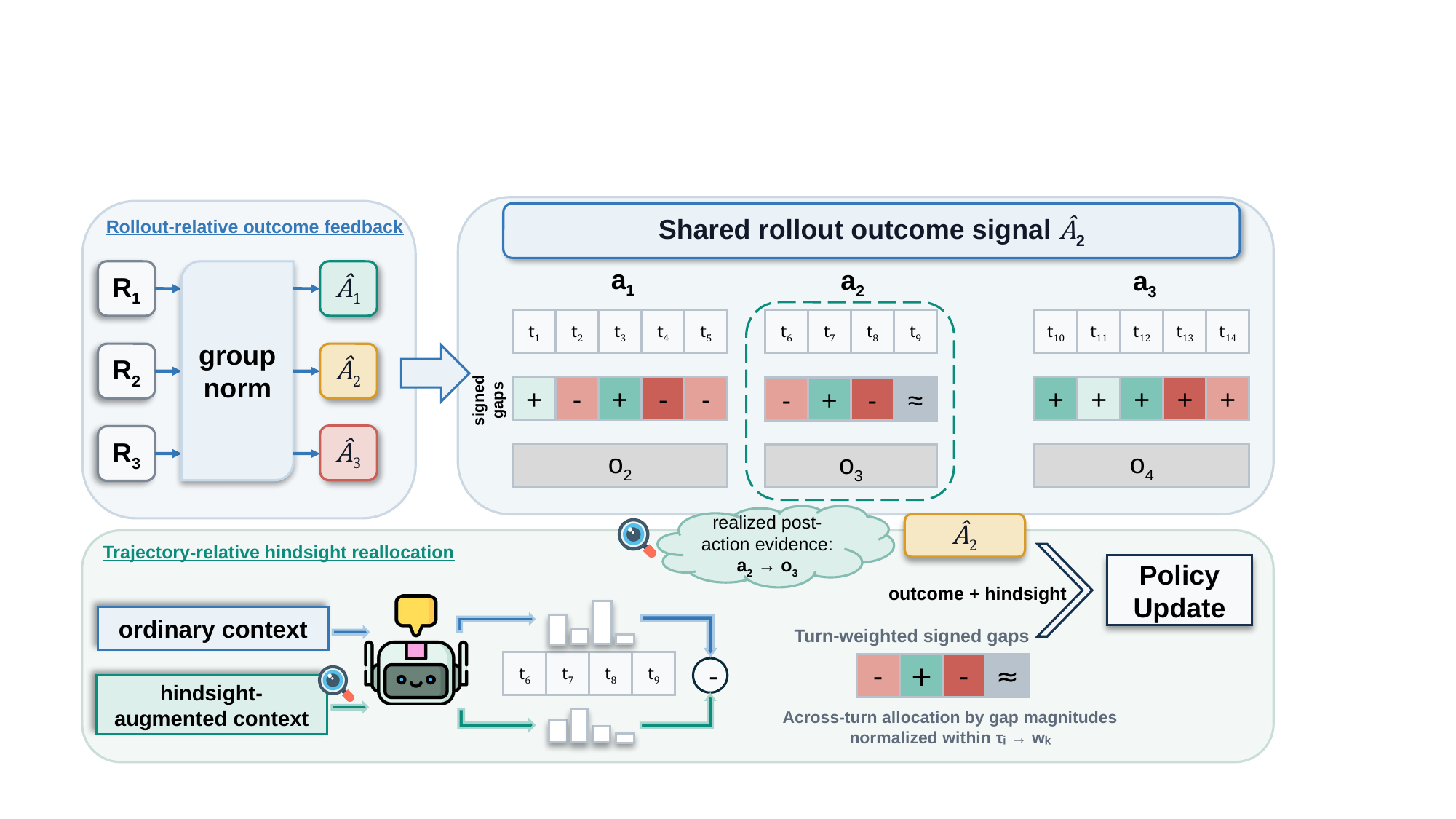}
  \caption{\method{} augments GRPO on the same realized trajectory. GRPO's
  across-rollout comparison yields one outcome signal $\widehat A_i$ shared by
  all action tokens in rollout $i$. A frozen training-time scoring view
  evaluates realized tokens with hindsight derived from action--observation
  consequences;
  signed gaps determine token direction and local strength, while a normalized
  turn profile determines relative allocation. The aligned signals form one token update field,
  and deployment retains only the ordinary policy.}
  \label{fig:method-overview}
\end{figure*}

\method{} complements GRPO's comparison across rollouts with a relative hindsight
update within each rollout. The completed interaction provides a
training-only policy view that reveals how the probability of each realized
token changes under hindsight. Each signed gap supplies token-level update
direction and local revision strength. \method{} aggregates the absolute gaps into
a trajectory-normalized profile that reallocates this dense signal across
turns. Figure~\ref{fig:method-overview} summarizes the resulting update.

\subsection{Preliminaries}

For a task $x$, the rollout policy $\pi_{\mathrm{old}}$ samples a group of $G$
multi-turn trajectories. A trajectory
$\tau=(o_1,a_1,o_2,\ldots,a_K,o_{K+1})$ contains each response
$a_k=(y_{k,1},\ldots,y_{k,T_k})$ together with its preceding observation
$o_k$ and realized post-action observation $o_{k+1}$. Given the final
environment return
$R_i$ of rollout $i$, GRPO computes
\begin{equation}
  \widehat A_i=\frac{R_i-\overline R}{\sigma_R+\epsilon_A},
  \label{eq:grpo-advantage}
\end{equation}
where $\overline R$ and $\sigma_R$ are the mean and standard deviation within
the rollout group. GRPO then optimizes its standard clipped surrogate loss
$\mathcal L_{\mathrm{GRPO}}$~\citep{schulman2017proximal,shao2024deepseekmathpushinglimitsmathematical}.
\method{} leaves this advantage and the GRPO clipping rule unchanged.

At token $t$ of turn $k$, $h_{k,t}$ denotes the ordinary interaction history,
including the response prefix $y_{k,<t}$. A binary mask $m_{k,t}$ marks tokens
eligible for hindsight learning. We use
$n_k=\sum_{t=1}^{T_k}m_{k,t}$ for the eligible tokens in turn $k$ and
$N_\tau=\sum_{k=1}^{K}n_k$ for the trajectory total. In our experiments,
all generated response tokens are eligible, including reasoning tokens when
they are present in the generated response; prompt and padding tokens are
excluded.\looseness=-1

\subsection{Trajectory-Relative Hindsight Allocation}

\paragraph{Hindsight-conditioned policy gap.}
Once a realized trajectory is available, the same hindsight protocol is
applied at every decision turn. It constructs the turn-aligned outcome view
\begin{equation}
  z_k=\operatorname{OutcomeView}(\tau,k),
  \label{eq:outcome-view}
\end{equation}
which exposes the locally realized consequence of the decision at turn $k$.
An environment interface maps native interaction records into this shared
turn-aligned object. In our experiments, the frozen training-time scoring
snapshot is the pre-update rollout snapshot, $\pi_T=\pi_{\mathrm{old}}$, so the
profile-building gap $\Delta^{\mathrm{old}}$ isolates context augmentation.
The scoring snapshot is a training-time policy snapshot rather than a
separately trained teacher, and the realized response being scored is
unchanged.
Conceptually, the two token contexts are
\begin{equation}
  \begin{aligned}
    h_{k,t}&=[P_k;\,y_{k,<t}],\\
    P_k^+&=\operatorname{Aug}(P_k,z_k),\\
    h^+_{k,t}&=[P_k^+;\,y_{k,<t}],
  \end{aligned}
  \label{eq:context-layout}
\end{equation}
where $P_k$ is the ordinary prompt at turn $k$. The realized trajectory
provides one outcome view for the turn; the recorded response and its token
prefix are otherwise unchanged. Appendix~\ref{sec:appendix-hindsight} details
the shared protocol and its serialization. For any policy
$\pi$ evaluated on the ordinary history, let $\operatorname{clip}_b(u)$ clip
$u$ to $[-b,b]$ and define
\begin{equation}
  \Delta_{k,t}(\pi)=\operatorname{sg}\!\left[
  \operatorname{clip}_c\!\left(
  \log\frac{\pi_T(y_{k,t}\mid h^+_{k,t})}
  {\pi(y_{k,t}\mid h_{k,t})}\right)\right],
  \label{eq:policy-gap}
\end{equation}
where $\operatorname{sg}$ denotes stop-gradient. Here $\operatorname{clip}_c$
is an elementwise bound on the log-probability gap, while the outer clamp in
Eq.~\ref{eq:joint-objective} bounds the scalar dense contribution. We write
$\Delta^{\mathrm{old}}_{k,t}=\Delta_{k,t}(\pi_{\mathrm{old}})$ when building
the allocation profile and
$\Delta^\theta_{k,t}=\Delta_{k,t}(\pi_\theta)$ during actor optimization.
Equation~\ref{eq:policy-gap} follows on-policy self-distillation: the frozen
scoring view has access to hindsight, while the target remains the student's
own sample
\citep{agarwal2024onpolicydistillationlanguagemodels,
zhao2026selfdistilledreasoneronpolicyselfdistillation}.

\paragraph{Turn score and relative allocation.}
For every turn containing at least one eligible token, \method{} summarizes the
absolute hindsight gap:
\begin{equation}
  s_k=\frac{1}{n_k}\sum_{t=1}^{T_k}
  m_{k,t}\left\lvert\Delta^{\mathrm{old}}_{k,t}\right\rvert.
  \label{eq:turn-score}
\end{equation}
Turns with $n_k=0$ are omitted from the profile and receive no dense update.
The token-weighted trajectory average removes the common gap scale. The
remaining ratio is the \method{} turn weight:
\begin{equation}
  \begin{gathered}
    \overline s_\tau=\frac{1}{N_\tau}
    \sum_{j:n_j>0}n_js_j,\qquad
    w_k=\frac{s_k}{\overline s_\tau},\\
    \sum_{k:n_k>0}n_kw_k=N_\tau.
  \end{gathered}
  \label{eq:turn-weight}
\end{equation}
The last identity gives the weights an eligible-token-weighted mean of one.
Equivalently, $w_k$ compares turn $k$'s share of the absolute hindsight gap
with its share of eligible tokens. A weight above one therefore identifies a
turn whose revision is larger than its token count alone would predict. If
the total gap is numerically zero, \method{} uses $w_k=1$; a single eligible turn
also reduces to unit weighting. The profile is invariant to a common positive
rescaling of all token gaps in the trajectory.

This normalization makes the allocation comparable to a unit-profile control:
both expose the eligible tokens to a mean multiplier of one, but \method{}
redistributes that multiplier across turns. The score $s_k$ measures the
magnitude of a hindsight-conditioned policy discrepancy, not the causal
importance of turn $k$ to the final
reward. For example, if one turn changes substantially under hindsight while
two equally long turns barely change, \method{} concentrates the dense update
on the first turn without changing the trajectory's mean multiplier.
For three equally long turns with $s=(0.1,0.2,0.3)$, for example,
$\overline s_\tau=0.2$ and $w=(0.5,1,1.5)$.
The mean-one identity calibrates allocation rather than loss magnitude: signed
token gaps and token losses still determine the realized dense update.

\subsection{Joint Optimization}

\paragraph{Dense signed hindsight objective.}
During actor optimization, \method{} refreshes the detached gap
$\Delta^\theta_{\tau,k,t}$ against the current policy. Let
$M=\sum_{\tau,k,t}m_{\tau,k,t}$ be the number of eligible tokens in the actor
microbatch. The dense hindsight objective is the masked token mean
\begin{equation}
  \begin{aligned}
    \mathcal L_{\mathrm{dense}}
    &=-\frac{1}{M}\sum_{\tau,k,t}
    m_{\tau,k,t}w_{\tau,k}\Delta^\theta_{\tau,k,t}\,\\
    &\qquad\times
    \log\pi_\theta(y_{\tau,k,t}\mid h_{\tau,k,t}).
  \end{aligned}
  \label{eq:dense-loss}
\end{equation}
We set this loss to zero when the microbatch contains no eligible tokens. A
positive gap increases the probability of the realized token, whereas a
negative gap decreases it; its absolute value sets the local strength before
turn reweighting.
Thus, Eq.~\ref{eq:turn-weight} controls where the dense update is
concentrated, while Eq.~\ref{eq:policy-gap} preserves its signed token-level
direction and local strength.

\paragraph{Joint objective.}
Let $\mathcal L_{\mathrm{out}}$ denote the sparse outcome-driven policy
objective, including $\mathcal L_{\mathrm{GRPO}}$ and any fixed regularization.
At update $g$, \method{} optimizes
\begin{equation}
  \begin{aligned}
    b_g&=\alpha\operatorname{sg}\!\left(
    \left\lvert\mathcal L_{\mathrm{GRPO}}\right\rvert\right),\\
    \mathcal L(\theta)
    &=\mathcal L_{\mathrm{out}}(\theta)
    +\operatorname{clip}_{b_g}\!\left(
    \lambda_g\mathcal L_{\mathrm{dense}}\right).
  \end{aligned}
  \label{eq:joint-objective}
\end{equation}
The coefficient $\lambda_g$ is zero during warmup and equals the configured
dense-feedback coefficient afterward. Applied once to the actor-microbatch
scalar, the clamp bounds the hindsight contribution relative to the detached
GRPO loss magnitude. When the dense term
saturates either boundary, its gradient is zero and the outcome-policy gradient
remains active.

The allocation profile is constructed from the frozen scoring snapshot and
detached before actor optimization. The actor therefore receives a fixed
trajectory-level allocation target while refreshing only the signed token gap
in the current policy; this separates where supervision is assigned from the
policy update that applies it.

Algorithm~\ref{alg:training-update} summarizes the shared optimization procedure.
The interaction interface implements only $\operatorname{OutcomeView}$ and
$\operatorname{Aug}$, translating native interaction records into the same
turn-aligned semantic view. The semantic scoring, trajectory normalization,
and optimization rule are shared across environments; implementation schedules
and resource settings are specified in Appendix~A.

\begin{algorithm}[H]
\caption{\method{} training update}
\label{alg:training-update}
\small
\textbf{Input:} actor $\pi_\theta$, task batch $\{x_b\}$, interaction interface
$(\operatorname{OutcomeView},\operatorname{Aug})$.
\begin{enumerate}[leftmargin=1.45em,label=\arabic*.,itemsep=1pt,topsep=2pt,
  parsep=0pt,partopsep=0pt]
  \item Freeze the pre-update snapshot $\pi_{\mathrm{old}}$ and use it as the
  scoring snapshot $\pi_T$ while collecting groups of completed on-policy
  trajectories $\{(\tau_i,R_i)\}$.
  \item Compute $\widehat A_i$ and the outcome objective
  $\mathcal L_{\mathrm{out}}$ from group returns.
  \item For each turn $k$, extract
  $z_{i,k}=\operatorname{OutcomeView}(\tau_i,k)$ and form
  $P^+_{i,k}=\operatorname{Aug}(P_{i,k},z_{i,k})$ and
  $h^+_{i,k,t}=[P^+_{i,k};y_{i,k,<t}]$.
  \item Score the same realized tokens under ordinary and augmented contexts;
  compute $\Delta^{\mathrm{old}}_{i,k,t}$, $s_{i,k}$, and the detached
  trajectory-normalized profile $w_{i,k}$.
  \item During actor optimization, refresh $\Delta^\theta_{i,k,t}$, construct
  $\mathcal L_{\mathrm{dense}}$, and update $\theta$ with
  Eq.~\ref{eq:joint-objective}.
  \item Discard the hindsight evidence, augmented contexts, and profile.
\end{enumerate}
\textbf{Output:} updated ordinary policy $\pi_\theta$.
\end{algorithm}

Deployment uses only $\pi_\theta$ with the ordinary interaction history and
requires no hindsight module, teacher, or routing procedure.

\section{Experiments}
\label{sec:experiments}

\subsection{Experimental Setup}

\paragraph{Benchmarks, models, and evaluation.}
We evaluate Qwen2.5-3B-Instruct~\citep{qwen2024qwen25technicalreport} and
Qwen3-1.7B~\citep{yang2025qwen3technicalreport} on WebShop~\citep{yao2022webshop}
and ALFWorld~\citep{shridhar2021alfworldaligningtextembodied}. Training uses
online rollouts from each environment's official training split;
the task batch supplies scheduled task instances, while interaction histories,
rewards, and hindsight signals are obtained from the live environment
interaction. WebShop reports
success and dense task score on the complete official test split of 500 goals.
ALFWorld reports success on every game in the official Seen and Unseen
evaluation splits (140 and 134 games, respectively), keeping the two splits
separate. Training-time validation uses a fixed 128-instance diagnostic subset
and is distinct from the final evaluation procedure. Every reported method is
finally run with the same complete-set evaluator and benchmark metric
implementation.

\paragraph{Baselines.}
The comparison set includes same-stack GRPO~\citep{shao2024deepseekmathpushinglimitsmathematical}
and implementations of SERL~\citep{li2026distillselectivehindsightdistillation},
SDAR~\citep{lu2026selfdistilledagenticreinforcementlearning},
GRPO+OPSD~\citep{zhao2026selfdistilledreasoneronpolicyselfdistillation}, and
RLSD~\citep{yang2026selfdistilledrlvr}. We preserve the released methods'
native objectives and recommended settings rather than rewriting them as variants
of \method{}.

\paragraph{Unified hindsight protocol.}
Both environments use the protocol in Eq.~\ref{eq:outcome-view}: each decision
turn is paired with a training-only outcome view of its locally realized
consequence, and the same recorded response is scored with and without that
view. The interaction interface maps native event records into this shared
turn-aligned view; the same-response scoring rule, trajectory-level
normalization, and deployment interface are shared. The outcome-view pathway
is held fixed among Uniform, Permuted, and \method{} within each controlled
comparison, so these experiments vary the allocation profile rather than the
hindsight construction.

\paragraph{Controlled-study configuration.}
The profile study compares Uniform, Permuted, and \method{} with Qwen3-1.7B on both
environments; GRPO provides the outcome-only reference. Within each
environment, the three hindsight configurations share batch, rollout,
trajectory-length, and optimization settings. All systems train for 150
optimization steps, and group-relative methods sample eight trajectories per
task instance. Main-body dynamics use Qwen3-1.7B ALFWorld, whose longer
trajectories make turn allocation easiest to inspect. The appendix specifies
the hindsight serialization, full hyperparameters, and hardware.

\subsection{Overall Agentic Performance}

\begin{table*}[t]
  \centering
  \small
  \setlength{\tabcolsep}{2.4pt}
  \renewcommand{\arraystretch}{1.02}
  \resizebox{\textwidth}{!}{%
  \begin{tabular}{@{}lrrrrrrr@{\hspace{4pt}}rrrrrrr@{\hspace{4pt}}rr@{}}
    \toprule
    & \multicolumn{14}{c}{ALFWorld success} & \multicolumn{2}{c}{WebShop} \\
    \cmidrule(lr){2-15}\cmidrule(lr){16-17}
    Method & \multicolumn{7}{c}{Seen} & \multicolumn{7}{c}{Unseen} & Succ. & Score \\
    \cmidrule(lr){2-8}\cmidrule(lr){9-15}\cmidrule(lr){16-17}
    & P\&P & Two & Look & Heat & Cool & Clean & Avg. & P\&P & Two & Look & Heat & Cool & Clean & Avg. & & \\
    \midrule
    \multicolumn{17}{l}{\textit{Qwen2.5-3B}} \\
    GRPO       & 80.0 & 41.7 & \underline{61.5} & 50.0 & 44.0 & 63.0 & 58.6 & 66.7 & 29.4 & 55.6 & 52.2 & 52.4 & 45.2 & 50.7 & 66.2 & 81.8 \\
    SERL       & \textbf{97.1} & \textbf{66.7} & 53.8 & \textbf{81.2} & 76.0 & \underline{81.5} & \textbf{79.3} & \underline{79.2} & \textbf{82.4} & 33.3 & 69.6 & 66.7 & \underline{77.4} & 69.4 & 68.6 & 82.2 \\
    SDAR       & 91.4 & 50.0 & 53.8 & \underline{75.0} & \textbf{84.0} & \textbf{88.9} & \underline{77.1} & 75.0 & \underline{76.5} & \underline{61.1} & \underline{73.9} & \textbf{85.7} & 71.0 & \textbf{73.9} & \underline{69.0} & \textbf{82.6} \\
    GRPO+OPSD  & \underline{94.3} & 54.2 & \underline{61.5} & 56.2 & \textbf{84.0} & 77.8 & 75.0 & \textbf{83.3} & 58.8 & \underline{61.1} & 65.2 & 61.9 & \textbf{80.6} & 70.1 & 68.4 & 81.4 \\
    RLSD       & \underline{94.3} & 54.2 & \textbf{69.2} & \underline{75.0} & 72.0 & \underline{81.5} & 76.4 & 75.0 & 52.9 & 55.6 & \textbf{91.3} & 71.4 & 71.0 & 70.9 & 68.8 & 81.6 \\
    \textbf{\method{}} & \textbf{97.1} & \underline{58.3} & \textbf{69.2} & 62.5 & \underline{80.0} & \textbf{88.9} & \textbf{79.3} & \underline{79.2} & 52.9 & \textbf{77.8} & 69.6 & \underline{76.2} & \underline{77.4} & \underline{73.1} & \textbf{69.4} & \underline{82.4} \\
    \midrule
    \multicolumn{17}{l}{\textit{Qwen3-1.7B}} \\
    GRPO       & \textbf{80.0} & 33.3 & \underline{61.5} & \textbf{68.8} & \underline{64.0} & 63.0 & 62.9 & \underline{70.8} & 23.5 & \underline{50.0} & \underline{69.6} & \underline{61.9} & 61.3 & 58.2 & 56.4 & 78.7 \\
    SERL       & 71.4 & \textbf{45.8} & 46.2 & \textbf{68.8} & 60.0 & 70.4 & 62.1 & 62.5 & \underline{47.1} & \textbf{55.6} & 56.5 & 57.1 & 61.3 & 57.5 & 67.6 & 81.1 \\
    SDAR       & 74.3 & \textbf{45.8} & \textbf{69.2} & \underline{62.5} & 60.0 & \underline{74.1} & \underline{65.0} & \textbf{75.0} & 35.3 & 44.4 & \underline{69.6} & \textbf{66.7} & 71.0 & \underline{62.7} & \underline{70.0} & \underline{81.8} \\
    GRPO+OPSD  & 68.6 & 37.5 & 53.8 & 56.2 & \underline{64.0} & 63.0 & 58.6 & 58.3 & 41.2 & \textbf{55.6} & 60.9 & 52.4 & 54.8 & 54.5 & 60.8 & 79.1 \\
    RLSD       & 71.4 & \underline{41.7} & 53.8 & 56.2 & 56.0 & \underline{74.1} & 60.7 & 62.5 & \underline{47.1} & 44.4 & 52.2 & 47.6 & \underline{74.2} & 56.7 & 68.4 & 81.4 \\
    \textbf{\method{}} & \underline{77.1} & \underline{41.7} & \underline{61.5} & \underline{62.5} & \textbf{68.0} & \textbf{88.9} & \textbf{68.6} & 66.7 & \textbf{64.7} & \underline{50.0} & \textbf{82.6} & \underline{61.9} & \textbf{77.4} & \textbf{68.7} & \textbf{75.2} & \textbf{85.7} \\
    \bottomrule
  \end{tabular}
  }
  \caption{Complete-set results (\%). ALFWorld is decomposed by task family;
  Avg. is split-level success rather than an unweighted family mean.
  WebShop reports success (Succ.) and task score. P\&P denotes pick-and-place;
  Two denotes placing two objects. Bold and underlining mark the best and
  second-best result per backbone and column, respectively.}
  \label{tab:main-results}
\end{table*}

Table~\ref{tab:main-results} reports the complete comparison in one matrix,
including all six ALFWorld task families rather than only split-level
averages. Across the eight aggregate environment metrics (ALFWorld Seen and
Unseen averages plus the two WebShop metrics for each backbone), \method{} is
higher than GRPO in every case and records or matches the highest point estimate
among all six methods in six. The pattern is strongest with Qwen3-1.7B, where \method{}
leads both WebShop metrics and both ALFWorld split averages. With Qwen2.5-3B,
\method{} also gives the highest WebShop success, ties the highest ALFWorld
Seen average, and improves all four aggregate metrics over GRPO.

The strongest gain appears with Qwen3-1.7B on WebShop. Relative to GRPO,
\method{} increases success from 56.4\% to 75.2\% and dense task score from
78.7\% to 85.7\%. The two metrics agree: the method completes more goals while
also improving partial task completion.

ALFWorld complements this outcome comparison with longer action sequences and
separate seen and unseen evaluations. On Qwen3-1.7B, \method{} exceeds the
strongest competing average by 3.6 points on Seen and 6.0 points on Unseen.
The category breakdown sharpens this result: different baselines lead
different task families, whereas \method{} attains the strongest overall
average on both splits without relying on a single dominant category. With
Qwen2.5-3B, it raises the Seen and Unseen averages over GRPO by 20.7 and 22.4
points, respectively, and ties the strongest Seen average. We use Qwen3-1.7B
ALFWorld for the main-body learning-dynamics analysis and both environments
for the controlled profile study.

WebShop's paired metrics show that the aggregate gain is not confined to
marginal goal completion. ALFWorld's family columns likewise show that the
split-level averages are not driven by one task type. Together, the benchmarks
evaluate exact and partial task completion alongside seen and unseen household
interaction rather than repeating one aggregate comparison.

\begin{figure*}[t]
    \centering
    \includegraphics[width=\textwidth]{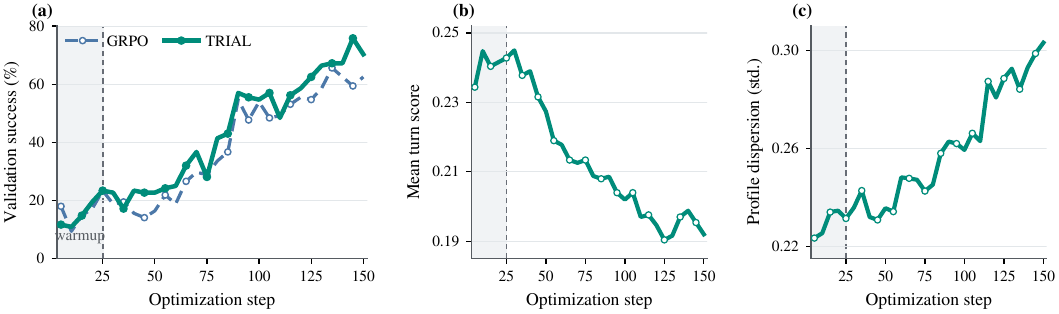}
    \caption{Qwen3-1.7B validation and allocation dynamics on ALFWorld.
    (a) Success on the fixed 128-game validation set, evaluated every five
    optimization steps. Along this diagnostic training run, \method{} exceeds GRPO
    at 22 of the 26 checkpoints after the dense term activates, and finishes
    at 70.3\% versus 62.5\%.
    (b--c) Mean absolute-gap turn score and standard deviation of the normalized
    turn profile, shown as non-overlapping five-step means on independent
    vertical scales. Shading marks steps 1--24 before the dense hindsight term
    activates; the dashed line marks step 25.}
    \label{fig:alfworld-dynamics}
\end{figure*}

\subsection{Controlled Revision Profiles}
\label{sec:analysis}

The controlled study compares GRPO, Uniform, Permuted, and the complete method
with Qwen3-1.7B on both WebShop and ALFWorld (Table~\ref{tab:coarse-ablation}).
Uniform, Permuted, and \method{} hold
the dense hindsight pathway fixed and vary only the applied turn profile;
GRPO is the outcome-only reference. Uniform uses unit multipliers. Permuted
assigns each active turn another active turn's multiplier within the same
trajectory, preserves relative multiplier ratios up to a common rescaling, and
restores a token-weighted mean of one.

\begin{table*}[t]
  \centering
  \small
  \setlength{\tabcolsep}{2.4pt}
  \renewcommand{\arraystretch}{1.02}
  \resizebox{\textwidth}{!}{%
  \begin{tabular}{@{}lrrrrrrr@{\hspace{4pt}}rrrrrrr@{\hspace{4pt}}rr@{}}
    \toprule
    & \multicolumn{14}{c}{ALFWorld success} & \multicolumn{2}{c}{WebShop} \\
    \cmidrule(lr){2-15}\cmidrule(lr){16-17}
    Method & \multicolumn{7}{c}{Seen} & \multicolumn{7}{c}{Unseen} & Succ. & Score \\
    \cmidrule(lr){2-8}\cmidrule(lr){9-15}\cmidrule(lr){16-17}
    & P\&P & Two & Look & Heat & Cool & Clean & Avg. & P\&P & Two & Look & Heat & Cool & Clean & Avg. & & \\
    \midrule
    GRPO & 80.0 & 33.3 & \textbf{61.5} & 68.8 & 64.0 & 63.0 & 62.9 & 70.8 & 23.5 & \textbf{50.0} & 69.6 & 61.9 & 61.3 & 58.2 & 56.4 & 78.7 \\
    Uniform & 71.4 & \textbf{45.8} & 38.5 & 62.5 & 48.0 & 77.8 & 60.0 & 75.0 & 41.2 & 33.3 & 60.9 & 61.9 & 61.3 & 57.5 & 62.8 & 77.8 \\
    Permuted & \textbf{82.9} & 33.3 & 53.8 & \textbf{75.0} & \textbf{68.0} & 74.1 & 66.4 & \textbf{79.2} & 58.8 & 33.3 & 65.2 & \textbf{71.4} & 67.7 & 64.2 & 56.8 & 76.3 \\
    \textbf{\method{}} & 77.1 & 41.7 & \textbf{61.5} & 62.5 & \textbf{68.0} & \textbf{88.9} & \textbf{68.6} & 66.7 & \textbf{64.7} & \textbf{50.0} & \textbf{82.6} & 61.9 & \textbf{77.4} & \textbf{68.7} & \textbf{75.2} & \textbf{85.7} \\
    \bottomrule
  \end{tabular}
  }
  \caption{Controlled revision-profile results with Qwen3-1.7B (\%). ALFWorld
  Avg. is split-level success; WebShop reports success (Succ.) and task score.
  P\&P denotes pick-and-place and Two placing two objects; bold marks each
  column's best result. The three hindsight variants share the outcome-view
  protocol and optimization: Uniform uses unit multipliers, Permuted shuffles
  source-turn assignments and restores a token-weighted mean of one, and
  \method{} uses the source-aligned relative profile. GRPO is the outcome-only
  reference.}
  \label{tab:coarse-ablation}
\end{table*}

Uniform provides the unit-weighted control for dense hindsight feedback.
Relative to GRPO,
it improves WebShop success by 6.4 points but does not yield consistent gains
on the other three metrics. Unit-weighted dense hindsight therefore does not
explain the full gain. Applying \method{}'s gap-dependent, source-aligned
relative profile raises all four metrics over Uniform: WebShop
success and score increase by 12.4 and 7.9 points, and ALFWorld Seen and Unseen
success increase by 8.6 and 11.2 points.

Permuted provides a stricter correspondence control. It preserves the
non-uniform multiplier values up to normalization but assigns them to
different source turns. Its effect varies by environment, remaining above
Uniform on ALFWorld while falling below it on WebShop. \method{} nevertheless
exceeds Permuted on all four aggregate metrics, by 18.4 and 9.4 points on
WebShop and by 2.2 and 4.5 points on ALFWorld. The controls therefore
distinguish dense hindsight, a non-uniform profile, and source-turn
correspondence. \method{}'s consistent lead supports applying revision to the turn
whose policy assessment changes, rather than merely adding dense feedback or
non-uniform weights.\looseness=-1

\subsection{Learning and Allocation Dynamics}

\begin{figure*}[t]
    \centering
    \includegraphics[width=\textwidth]{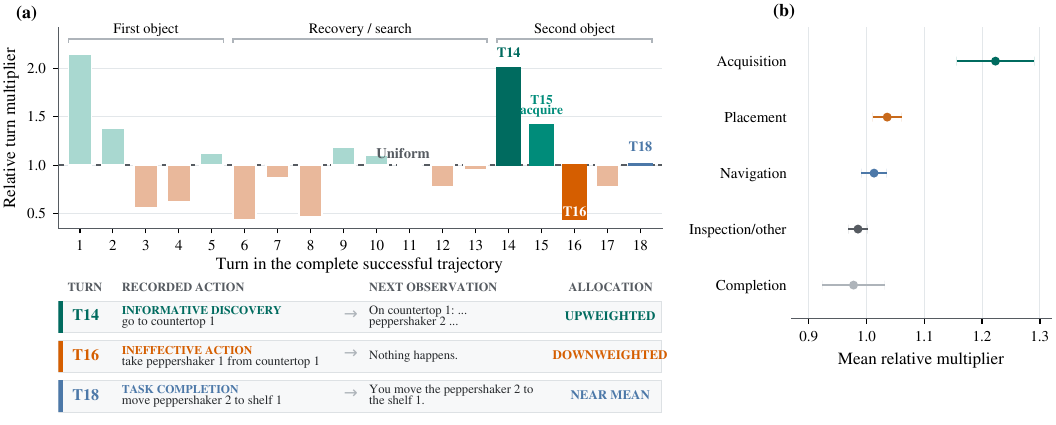}
    \caption{\method{}'s hindsight allocation on successful ALFWorld Pick-Two
    trajectories. (a) An 18-turn profile and three example decisions; T15 is
    the acquisition following T14's discovery. (b) Mean multipliers by decision
    class across 21 successful trajectories from five task instances.
    Acquisition is largest and completion remains near unit; error bars show
    SEM over trajectory-level class means.}
    \label{fig:allocation-trace}
\end{figure*}

\paragraph{Learning dynamics.}
We use Qwen3-1.7B ALFWorld as the representative long-horizon setting. The
three panels report success on the fixed validation set, the mean absolute-gap
turn score, and dispersion of the constructed trajectory-relative
turn-weight profile. Validation is evaluated every five optimization steps;
the two mechanism diagnostics are non-overlapping five-step means. After the
dense hindsight objective activates, \method{} exceeds GRPO at 22 of the 26
diagnostic checkpoints along this run and ends 7.8 percentage points higher.
The warmup boundary is marked explicitly.

\paragraph{Profile evolution.}
The diagnostic panels expose a complementary pattern. The mean turn score
decreases from roughly $0.23$--$0.24$ early in training to $0.19$
at the end, while the standard deviation of the normalized profile rises from
roughly $0.22$--$0.23$ to $0.30$. In other words, the absolute hindsight gap
contracts as training progresses, yet its relative distribution across turns
becomes more heterogeneous. This joint trend matches \method{}'s distinction between
local gap strength and relative allocation across turns.

\paragraph{Trajectory-level allocation.}
Figure~\ref{fig:allocation-trace} makes this heterogeneity concrete. In the
illustrative Pick-Two success, \method{} emphasizes both object-location discovery and
object-acquisition decisions. Several ineffective or redundant interactions
fall below the trajectory mean, including attempted acquisitions that produce
no state change. The pattern is not dominated by terminal turns:
acquisition has the largest mean multiplier across 21 successful trajectories,
while completion remains near the unit profile. \method{} thus concentrates
hindsight-derived supervision where the turn-aligned outcome view most changes
the policy's
assessment of the recorded decision, rather than treating every turn alike.
Together, the complete-set results establish outcome-level gains, the profile
controls test source-aligned allocation, and the diagnostics trace the same
mechanism through training and realized trajectories without additional
task-specific reward shaping or inference-time components.

\section{Conclusion}

\method{} combines rollout-level outcome comparison with turn-aligned hindsight:
signed local revisions form a mean-one trajectory-relative profile that
reallocates supervision while preserving outcome advantages and the average
multiplier. Across two environments and backbones, gains over GRPO, controls,
and diagnostics support emphasizing decisions most revised by the outcome view.

\section{Limitations}

Our evidence currently covers text-based interactive environments with
discrete actions. \method{} also assumes that a completed trajectory exposes
serializable post-action evidence that can be aligned with the decision that
produced it. Extending this interface to less structured interaction remains
future work. The hindsight view is confined to training; deployment retains the
ordinary policy interface without an additional model, hindsight input, or
routing procedure.
Each update also requires a hindsight-conditioned forward pass over generated
responses, increasing training cost relative to outcome-only GRPO. Our reported
runs use a single seed; cross-seed variance and statistical robustness remain
unmeasured.

\FloatBarrier
\bibliography{references}

\clearpage
\appendix
\section{Experimental Protocol}
\label{sec:appendix-protocol}

This appendix records the protocol needed to interpret and reproduce the
reported comparisons. It follows the order of the experimental argument:
environment and metric definitions, optimization settings, construction of the
training-only hindsight view, baseline implementation, and additional result
breakdowns.

\subsection{Environments and Metrics}

\paragraph{WebShop.}
WebShop~\citep{yao2022webshop} presents a shopping instruction and requires the
agent to search, navigate product pages, choose options, and purchase an item.
We cap interaction at 15 environment turns. Success is the percentage of goals
completed exactly; task score gives partial credit for satisfying the requested
attributes. Final training rollouts sample from the released training goal pool,
while final results use the complete official test split of 500 evaluation goals.

\paragraph{ALFWorld.}
ALFWorld~\citep{shridhar2021alfworldaligningtextembodied} presents a household
goal in a text-rendered embodied environment. We cap interaction at 50 turns and
report binary success separately on all 140 Seen and 134 Unseen evaluation games.
The six task families are Pick and Place, Pick Two and Place, Look at Object, Heat
and Place, Cool and Place, and Clean and Place. Training rollouts use the official
ALFWorld training split. We do not average the Seen and Unseen splits.

\paragraph{Validation and final evaluation.}
Training-time validation uses a fixed 128-instance diagnostic subset in each
environment. The ALFWorld curve in Figure~\ref{fig:alfworld-dynamics} evaluates
this same subset every five optimization steps. Main-table values instead use
the complete official evaluation sets above under the shared full-set
evaluator. Neither the 128-instance diagnostic runs nor the
complete evaluation runs are used to construct training updates.

\subsection{Training Configuration}

All systems are trained for 150 optimization steps. Group-relative methods
sample eight trajectories for each task instance, and all backbones use their
standard instruct templates. We use a learning rate of $1\times10^{-6}$.
Table~\ref{tab:training-config} summarizes the task-dependent settings shared
by GRPO and \method{}.

\begin{table}[t]
  \centering
  \small
  \begin{tabular}{lcc}
    \toprule
    Configuration & WebShop & ALFWorld \\
    \midrule
    Training task batch & 16 & 16 \\
    Rollouts per task & 8 & 8 \\
    PPO minibatch & 64 & 256 \\
    Microbatch per GPU & 8 & 32 \\
    Maximum prompt tokens & 4096 & 2048 \\
    Maximum response tokens & 512 & 512 \\
    Auxiliary coefficient & 0.005 & 0.01 \\
    Gap clipping bound $c$ & 2.0 & 2.0 \\
    Auxiliary activation step & 25 & 25 \\
    Relative clamp $\alpha$ & 1.0 & 1.0 \\
    \bottomrule
  \end{tabular}
  \caption{Training and \method{} configurations. The learning rate is
  $1\times10^{-6}$ and training lasts 150 optimization steps in both
  environments.}
  \label{tab:training-config}
\end{table}

The \method{} auxiliary coefficient is zero for steps 1--24 and takes the value in
Table~\ref{tab:training-config} from step 25 onward. The auxiliary loss operates
on generated response tokens. We clip each detached log-probability gap to
$[-2,2]$ before constructing turn scores and set the scalar auxiliary-loss
bound to the magnitude of the detached GRPO loss ($\alpha=1$).

The reported main \method{} runs and controlled profile studies set the frozen
scoring snapshot to the pre-update rollout policy,
$\pi_T=\pi_{\mathrm{old}}$. Within each controlled comparison, Uniform, Permuted,
and \method{} therefore use the same scoring parameters; the compared
intervention is the applied turn profile. Thus, $\Delta^{\mathrm{old}}$
compares the same parameters under ordinary and outcome-view contexts, while
the realized response and deployment interface remain unchanged.

\subsection{Training-Only Hindsight Views}
\label{sec:appendix-hindsight}

The method uses one semantic protocol in both environments. For every decision
turn, $\operatorname{OutcomeView}(\tau,k)$ recovers a turn-aligned local view
anchored at the decision's realized action--observation consequence and
serializes it into an augmented scoring context. The realized response tokens
remain unchanged, so
the ordinary policy and hindsight-conditioned view score the same token
sequence. This exact-token correspondence is required by
Eq.~\ref{eq:policy-gap}.

For turn $k$, let $P_k$ denote the ordinary prompt before the response. The
ordinary scorer receives $[P_k;y_{k,<t}]$, whereas the hindsight view receives
$[P_k^+;y_{k,<t}]$ with $P_k^+=\operatorname{Aug}(P_k,z_k)$. Thus, one outcome
view $z_k$ is shared by all scored tokens of the turn, and the complete
recorded response, including its action tokens, is unchanged between views.
The outcome view is constructed only after the interaction has produced the
corresponding consequence and is discarded after the update.

The concrete outcome view is aligned with the current decision. In WebShop,
it contains the page state reached by the current action; in ALFWorld, it
contains the text-rendered post-action state. Terminal turns use the available
terminal consequence. These are environment-specific
serializations of the shared $\operatorname{OutcomeView}$ interface rather
than different scoring protocols, and all outcome views are discarded after
optimization.

\subsection{Hindsight Context Examples}

The following schematic examples abbreviate the task prompt and history while
preserving the semantic placement of the outcome view used during training.
In both environments, the text after \textsc{Recorded response} is copied
unchanged into the ordinary and hindsight views. Line breaks inside the
schematic responses below are typographic only.

\paragraph{ALFWorld.}
Suppose the task is to place an apple in a target receptacle and the recorded
turn takes the apple from a cabinet. The two scoring sequences have the form
\begin{quote}
\small
\textsc{Ordinary view:}\quad
[task; interaction history; current observation]\\
\textsc{Recorded response:}\\
\texttt{<think>...</think>}\\
\texttt{<action>take apple 1 from}\\
\hspace*{1em}\texttt{cabinet 1</action>}

\medskip
\textsc{Hindsight view:}\quad
[outcome-view-augmented scoring prompt]\\
\textsc{Post-action evidence:}\quad
``After this response, the environment returns: You pick up apple 1 from
cabinet 1.''\\
\textsc{Recorded response:}\quad [the identical response above]
\end{quote}
Here $z_k$ is the text-rendered state observed after the recorded action. The
hindsight view does not replace the action or generate a revised trajectory;
it only rescores the recorded response with its realized consequence visible.

\paragraph{WebShop.}
Suppose the task requests a navy shirt and the current turn opens a candidate
product. The corresponding context includes the immediate page reached by that
action:
\begin{quote}
\small
\textsc{Ordinary view:}\quad
[task; interaction history; current search page]\\
\textsc{Recorded response:}\\
\texttt{<think>...</think>}\\
\texttt{<action>click[product]</action>}

\medskip
\textsc{Hindsight view:}\quad
[outcome-view-augmented scoring prompt]\\
\textsc{Post-action evidence:}\quad
``The product page shows color options Navy and Black.''\\
\textsc{Recorded response:}\quad [the identical response above]
\end{quote}
Both examples instantiate the same protocol: a turn-aligned view of the local
realized consequence is inserted into the scoring context before the unchanged
response. The number of native event records needed to serialize that semantic
view is determined by the environment's interaction boundary.

\subsection{Baseline Implementation}

GRPO is implemented in the same Agentic RL stack as \method{} and uses the same
task interface, rollout budget, and final evaluation protocol. SERL is run from
its released implementation with immediate environment feedback, its native
action-level objective, and the task settings used for our comparison. SDAR,
GRPO+OPSD, and RLSD are run from the released SDAR implementation with their
method-specific objectives and recommended hyperparameters. We preserve each
method's optimization rule rather than rewriting all baselines as variants of
our loss. Every reported method is finally
evaluated through the same complete-set evaluator and metric implementation.

\subsection{Compute and Reproducibility}

Training runs use eight H20 GPUs. The training configurations and evaluation
protocol above specify the settings needed to interpret the reported results.

\section{Additional Analysis Details}
\label{sec:appendix-analysis}

\begin{table*}[t]
  \centering
  \footnotesize
  \setlength{\tabcolsep}{4pt}
  \begin{tabular}{@{}p{0.12\textwidth}p{0.405\textwidth}p{0.405\textwidth}@{}}
    \toprule
    Stage & GRPO inference trace & \method{} inference trace \\
    \midrule
    Goal and discovery
      & \textit{Reason:} ``Check the countertop for a credit card.''
        T1: go to countertop 1 $\rightarrow$ creditcard 3 is observed.
        T2: take creditcard 3 $\rightarrow$ acquisition succeeds.
        T3: go to sidetable 1 $\rightarrow$ creditcards 1 and 2 are observed.
      & \textit{Reason:} ``Check the target, then the nearby table.''
        T1: go to dresser 1 $\rightarrow$ no credit card is present.
        T2: go to sidetable 1 $\rightarrow$ creditcards 1 and 2 are observed. \\
    First placement
      & After moving between the sidetable and dresser, the policy claims both
        cards are in inventory.
        T6: move creditcard 1 to dresser 1 $\rightarrow$ ``Nothing happens.''
        $\cdots$
        T9: take creditcard 1 $\rightarrow$ acquisition succeeds.
        T12: move creditcard 1 to dresser 1 $\rightarrow$ placement succeeds.
      & T3: take creditcard 1 $\rightarrow$ acquisition succeeds.
        [one no-change turn]
        T5: go to dresser 1 $\rightarrow$ dresser is reached.
        T6: move creditcard 1 to dresser 1 $\rightarrow$ placement succeeds. \\
    Second-card recovery
      & The policy repeatedly reasons that creditcard 2 is at the dresser or
        countertop, despite observing it at the sidetable.
        T13--T49 contain repeated searches and failed remote acquisitions;
        T50 reaches the sidetable only after the turn budget is exhausted.
      & After three failed remote acquisitions, the policy returns to the last
        observed location.
        T10: go to sidetable 1 $\rightarrow$ creditcard 2 is observed.
        T11: take creditcard 2 $\rightarrow$ acquisition succeeds.
        T13--T14: go to dresser 1; move creditcard 2 $\rightarrow$ task completes. \\
    Outcome
      & \textbf{Failure at 50 turns}; 18 actions produce no state change.
      & \textbf{Success at 14 turns}; 5 actions produce no state change. \\
    \bottomrule
  \end{tabular}
  \caption{Compressed deterministic inference traces for the same ALFWorld
  task: put two credit cards in the dresser. Quoted reasoning is abbreviated
  from the generated response; $\cdots$ and bracketed text denote omitted
  turns, not reconstructed actions.}
  \label{tab:paired-inference-case}
\end{table*}

\subsection{Learning-Dynamics Construction}

Figure~\ref{fig:alfworld-dynamics} contains measured validation points rather
than an interpolation of final test results. Panel (a) records success on the
fixed 128-game ALFWorld validation set at steps $5,10,\ldots,150$. Panels
(b--c) partition the corresponding step-level mechanism logs into
non-overlapping five-step windows and report arithmetic means. These training
diagnostics are reported separately from the complete-set results in
Table~\ref{tab:main-results}.

At each step, the mean turn score and profile dispersion are computed over the
turn rows in that actor update, including both successful and unsuccessful
training rollouts. Each five-step window therefore summarizes five consecutive
updates rather than five independent training runs.

\subsection{Allocation-Trace Construction}

Figure~\ref{fig:allocation-trace} uses one native training rollout batch.
Panel (a) selects a readable 10--25-turn successful
Pick-Two trajectory containing acquisition, ineffective interaction, recovery,
and completion. Panel (b) includes all 21 successful Pick-Two trajectories for
the five task instances in that batch. Action classes are assigned from the
environment action string: \texttt{take} marks acquisition, \texttt{move} or
\texttt{put} marks placement, \texttt{go to} marks navigation, and the terminal
turn marks completion; remaining actions form inspection/other. Error bars are
descriptive SEM across trajectory-level category means and are not used for a
significance claim.

\subsection{Paired Inference Case Study}
\label{sec:appendix-case}

Table~\ref{tab:paired-inference-case} compares deterministic inference traces
from GRPO and \method{} on the same Seen Pick-Two game. Both policies receive
the same initial state and admissible-action interface. We retain the generated
reasoning cue when it explains the next decision, show the realized observation
after each displayed action, and mark omitted turns explicitly. The contrast is
behavioral: aggregate comparisons remain in Table~\ref{tab:main-results}, while
Figure~\ref{fig:allocation-trace} analyzes the training-time allocation
mechanism.

\end{document}